\documentclass[conference]{IEEEtran}
\usepackage{times}

\usepackage[numbers]{natbib}
\usepackage{multicol}
\usepackage[bookmarks=true]{hyperref}
\usepackage{multirow}

\usepackage[table,xcdraw]{xcolor}

\newif\ifpaperfinal
\paperfinaltrue

\ifpaperfinal
\newcommand{\iinote}[1]{}
\newcommand{\sdnote}[1]{}
\newcommand{\stnote}[1]{}
\newcommand{\tynote}[1]{}
\newcommand{\bknote}[1]{}
\newcommand{\nnote}[1]{}
\newcommand{\finaliinote}[1]{\textcolor{black}{#1}}
\else
\newcommand{\iinote}[1]{}
\newcommand{\finaliinote}[1]{\textcolor{orange}{\textbf{II: #1}}}
\newcommand{\tynote}[1]{\textcolor{red}{\textbf{Tian: #1}}}
\newcommand{\nnote}[1]{}
\newcommand{\stnote}[1]{\textcolor{blue}{\textbf{ST: #1}}}

\fi
\newcommand\numberthis{\addtocounter{equation}{1}\tag{\theequation}}
\usepackage{amsmath}
\usepackage{graphicx}
\usepackage{longtable,booktabs,array}
 \usepackage{multirow}

\begin{document}

\title{Improved Inference of Human Intent by Combining Plan Recognition and Language Feedback}

\author{
    Ifrah Idrees\textsuperscript{\rm 1}, 
    Tian Yun\textsuperscript{\rm 1}, 
    Naveen Sharma\textsuperscript{\rm 1},
    Yunxin Deng\textsuperscript{\rm 1}, 
    Nakul Gopalan\textsuperscript{\rm 2},
    George Konidaris\textsuperscript{\rm 1}, 
    Stefanie Tellex\textsuperscript{\rm 1} \\
    \textsuperscript{\rm 1}Brown University, USA, 
    \textsuperscript{\rm 2}Arizona State University, USA
}

\maketitle
\begin{figure*}[ht!]
  \centering
  \includegraphics[width=0.9\textwidth]{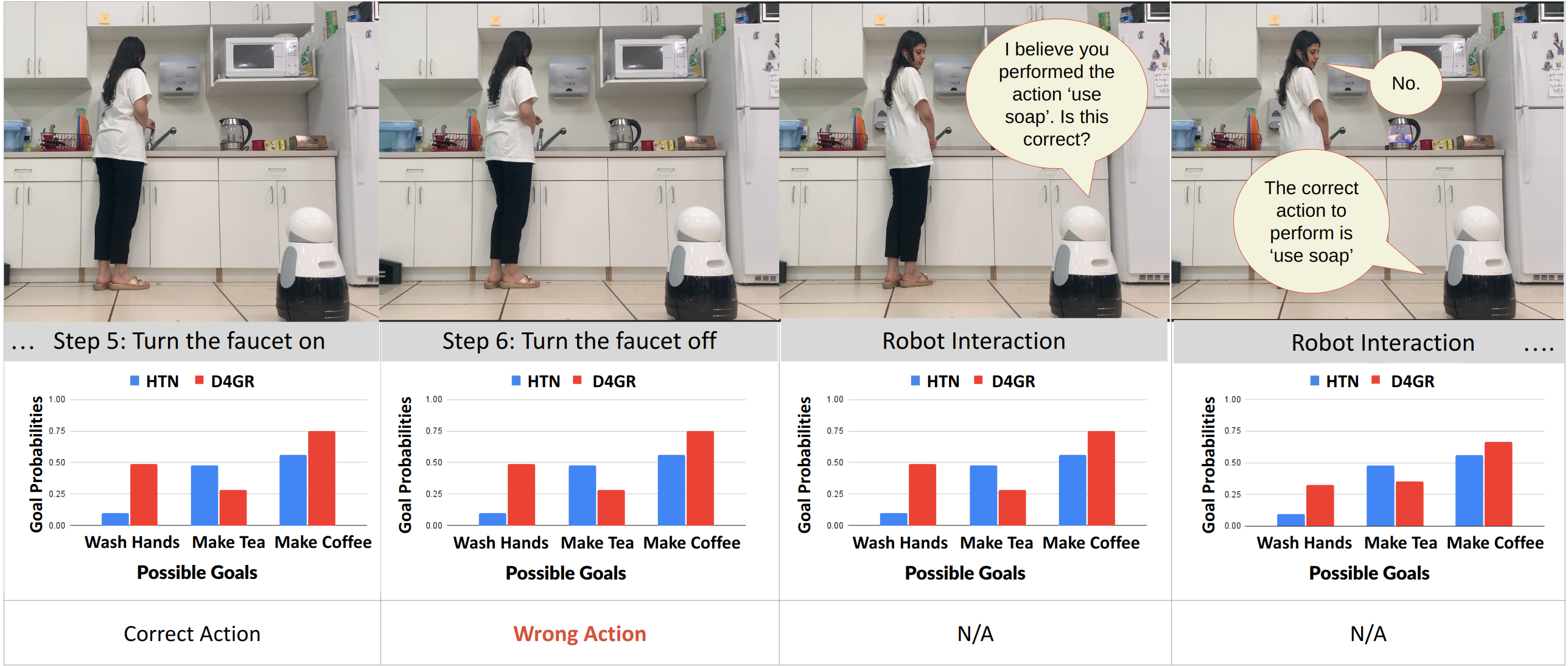}
  \caption{Kuri robot performing human intent (plan and goal) recognition. The human, while making coffee, starts a new goal of washing hands. However, the human forgets to use soap after turning on the faucet and instead turns off the faucet. The robot (with access to the Hierarchical Task Network representation of goals) observes that the current world observations do not progress the previous goal and also cannot lead to a new goal completion and hence uses D4GR to ask a clarification question.
  Based on  language feedback, the robot reduces its confidence in the goal wash hands and suggests the next action as using soap to complete one of the most plausible goals of washing hands.
  }
  \vspace{-2mm}
  \label{Fig-storyboard}
\end{figure*}

\begin{abstract}
Conversational assistive robots can aid people, especially those with cognitive impairments, to accomplish various tasks such as cooking meals, performing exercises, or operating machines.
However, to interact with people effectively, robots must recognize human plans and goals from noisy observations of human actions, even when the user acts sub-optimally. Previous works on Plan and Goal Recognition (PGR) as planning have used hierarchical task networks (HTN) to model the actor/human.
However, these techniques are insufficient as they do not have user engagement via natural modes of interaction such as language. Moreover, they have no mechanisms to let users, especially those with cognitive impairments, know of a deviation from their original plan or about any sub-optimal actions taken towards their goal.
We propose a novel framework for plan and goal recognition in partially observable domains---\textbf{D}ialogue for \textbf{G}oal \textbf{R}ecognition (\textbf{D4GR})   
enabling a robot to rectify its belief in human progress by asking clarification questions about noisy sensor data and sub-optimal human actions.
We evaluate the performance of D4GR over two simulated domains---kitchen and  blocks domain. With language feedback and the world state information in a hierarchical task model,
we show that D4GR framework for the highest sensor noise performs 1\% better than HTN in goal accuracy in both domains. For plan accuracy, D4GR outperforms by 4\% in the kitchen domain and 2\% in the blocks domain in comparison to HTN.
The ALWAYS-ASK oracle outperforms our policy by 3\% in goal recognition and 7\% in plan recognition.
D4GR does so by asking 68\%  fewer questions than an oracle baseline.  We also demonstrate a real-world robot scenario in the kitchen domain, validating the improved plan and goal recognition of D4GR in a realistic setting.
\end{abstract}
\IEEEpeerreviewmaketitle

\section{Introduction}


\nnote{Things to discuss: overall story novelty over related work?, experiment designs and results, Ask: simulation and robot demo only fine? or simulation or user study. Update: made Kuri work. }
\nnote{The first paragraph should generally describe the problem, focusing on the problem, what it is, and why it is important.}  
\nnote{talk about the first photo}
\nnote{They can do so by building situational awareness and building a rapport with their human partner.}

People with cognitive impairments, such as dementia, often struggle with focusing on everyday tasks and have limited attention spans.
Efforts to assist people in tracking task progress have involved various approaches, including modeling tasks as a hierarchical task network (HTN), using a Bayesian Hidden Markov Model, or employing a Partially Observable Markov Decision Process (POMDP) \cite{hoey2010automated,wang2017hierarchical}.
However, previous approaches focus on observing users rather than engaging them in an interaction. 
Our work aims to develop a model for a robot capable of assisting people complete tasks with language-based interactions; even when the users perform sub-optimal actions or switch between multiple goals. Our robot tracks the task progress using observations 
and question-asking using natural language.  
Such a robot can also benefit an operator building a machine, a child with autism doing their homework, or a child learning to do chores.

Inferring the goals and intents of the human requires plan and goal recognition (PGR) using noisy evidence from action execution that can be done efficiently using planning techniques \cite{meneguzzi2021survey}.
One key challenge in human intent recognition as a PGR problem is that the robot has partial observability of human intentions. This is compounded by noisy sensors that also create partial observability of the environment.
Modeling human progress during hierarchical tasks has been done using Hierarchical Task Networks (HTNs) \citep{wang2017hierarchical, holler2018plan}. However, these recognition techniques again do not engage with the users and 
assume that the user acts optimally. 
Moreover, incorporating clarification questions and language
utterance spoken by humans in PGR is challenging because of the huge space of language observations.
The existing solution to this problem is heuristics
\citep{context-awareness, fasola2013socially, kidd2008robots}, which
are prone to fail as the tasks and environment sensors become complex and noisy.

\nnote{The second paragraph should succinctly describe related approaches and why they do not solve the problem. }  
\nnote{Comment: In \citet{kidd2008robots}, dialogue grounded in environment but dialogue is hard coded. \citet{fasola2013socially} is heuristic based or hardcoded.}
\iinote{trick - computation or modeling}
\nnote{The next few paragraphs should describe our technical approach for solving the problem.}

\nnote{https://hierarchical-task.net/publications/hplan/2021/HPlan2021-paper9.pdf}
The main contribution of our paper is a novel formulation that combines expressive and hierarchical task representation of HTNs to represent the human mental states with the sequential decision-making capabilities of a Partially Observable Markov Decision Process, in our \textbf{D}ialogue for \textbf{G}oal \textbf{R}ecognition (\textbf{D4GR}). Our method keeps track of the environment, user state and dialogue history internally to perform PGR and guide in successful task completion.

POMDPs can model the uncertainty the robot faces as it performs intent recognitions and enables the robot to ask information-seeking questions.  However, 
POMDP planners traditionally do not decide the relevance of the state using the task network at each sequential time step. Further, the HTNs have no notion of rewards to generate a sequence of actions to maximize the agent's utility.
\nnote{IMP: We use the HTN  to represent the human mental state because we assume the person is a planner with goals and subgoals that are described hierarchically. Our robot agent is a MOMDP planner that performs belief updates given observations and performs long-horizon planning for asking clarification questions.}
\nnote{\stnote{Cut: 
Our dialogue manager builds context-awareness using:
\begin{itemize}
    \item An HTN, which represents ordered and unordered tasks, including subtasks and the sensors involved. This component maintains a belief over the human's mental state and progress in the task.
    \item A sequential decision-making representation---a POMDP that plans based on the dialogue intents and beliefs over human progress and world state to generate a policy for the dialogue manager.
\end{itemize}}}
To solve these challenges, we assume that the user is a planner with goals and subgoals that are represented hierarchically. 
Moreover, the robot is a POMDP planner performing long-horizon dialogue policy planning. This enables the robot to reason about asking meaningful questions in ambiguous settings, such as the user switching goals during multiple concurrent tasks and also performing sub-optimal actions. Using this information,  the robot can better recognize human intents based on the human actions estimated from noisy sensors and their language feedback. This model also explicitly allows for sub-optimal plans by a human user, which D4GR can detect.

   

\nnote{Remove: To achieve a framework for ADL task completion that intelligently asks questions as well as extracts implicit information as a human progresses towards completing the task, we define the Hierarchical Task Network(HTN) based Partially Observable Markov Decision Process, or DGR-POMDP. Our algorithm uses a POMDP planner to ask clarifying questions when the robot is confused and determines the desired goal/task that the human is trying to complete by using HTN-based belief update for interpreting natural language input as well as the state of partially observable environment sensors and using POMDP planner to ask clarifying questions when confused. Our model is able to understand the implicit meaning in the human's actions, known as implicatures/human intention. Implicatures are the inferences a listener makes when
bridging together a speaker's utterances and assumptions that
the speaker is acting cooperatively. For example, a person with dementia can do a wrong instruction step say  "turning off the faucet" instead of "turning on the faucet," that is needed to be completed to reach the goal of "washing hands" in such a scenario when DGR-POMDP asks the human about whether they did "turn on faucet," the elder's response will be no, and the implicature will be that some other action is performed.}

\nnote{Remove: The last paragraph should describe the evaluation and how we show that we have solved it.   I like to have a figure on the first page of the paper (Figure 1) that summarizes or shows what the robot can do that it couldn't do before.  This figure is often the first thing reviewers see and should show the "impressive result" of the paper visually, often a storyboard of the video.}

We evaluate the usefulness of D4GR by measuring the improved accuracy in PGR and comparing the planning time and number of questions asked to two
state-of-the-art baselines
in the simulated domain developed by \citet{wang2017hierarchical}.
Our system is able to more accurately infer human intents than these baselines using information gathered from a language without asking unnecessary questions. 
\iinote{Long-horizon planning enables our system to ask clarification questions when the situation is ambiguous such as user switching goals during multiple concurrent tasks and also performing sub-optimal actions.}
We run 880 \iinote{1000} trials for varying sensor noise levels where the simulated human tries to complete a combination of 
tasks in two domains - the kitchen and  blocks domain. In the kitchen domain, there are three tasks - washing hands, making a cup of tea, and making a cup of coffee while in the blocks domain, tasks include an assortment of stacking letters to make 4-7 lettered words - $rote$, $tone$, $tune$, $hawk$, $capstone$.  
\nnote{We also collect a real-world corpus of 5 pairs of human-human dialogue simulating a robot agent and a person with dementia trying to complete tasks for our pilot study that is used for setting our POMDP model.}  
\iinote{We find that our D4GR-POMDP planner can ask 44\% more questions than the no language state-of-the-art baseline while still maintaining 86\% high accuracy at estimating the state in all the trials. This is 66\% fewer questions and 1\% lesser accuracy than the baseline that always asks a correct  question. 
\nnote{Our robot demonstration video can be found here  \href{https://youtu.be/htwMtlsZnss}{robot demo}}.}
With language feedback and the world state information in a hierarchical task model, we show that our D4GR framework  outperforms HTN by 4\% on plan accuracy in the kitchen domain and 2\% in the blocks domain. In goal accuracy, for the highest sensor noise, our D4GR performs 1\% better than HTN in both kitchen and block domains. 
We also deployed our algorithm on a social robot Kuri as a demonstration of a socially intelligent robot helping confused users complete tasks. In this demonstration, Kuri performs improved PGR by asking clarification questions and reducing uncertainty for a challenging scenario where the user switches between multiple concurrent goals (e.g., washing hands and making coffee) and acts sub-optimally in the kitchen domain. 
The demo can be found here: \url{https://youtu.be/Om91zBiDDEY}. An example sequence of the user and robot interaction can be seen in Fig-\ref{Fig-storyboard}.  

\section{Related Work}

\iinote{Does the order make sense?}
\textbf{Plan and Goal Recognition as Planning:} The first work on PGR as planning was introduced by \citet{ramirez2009plan}. This research leverages classical planning systems to solve PGR problems. \citet{wang2017hierarchical} proposed an algorithm for PGR based on hierarchical task network~\citep{erol1994htn} that handles noisy sensors and sub-optimality in human actions. However, their approach detects mistakes heuristically using a manually defined threshold while performing a one-step look ahead without long-horizon planning. Additionally, it lacks dialog conversation to improve belief in the actor's progress in the task. Another relevant work that does not engage with the user is by \citet{zhi2020online}, which performs online Bayesian goal inference by modeling the agent as a boundedly rational planning agent but is not designed and evaluated for multiple concurrent hierarchical goals.
\citet{mirsky2017sequential} presents favorable results for the hypothesis that feedback from the acting agent can improve plan (goal and step) recognition, but their paper performs goal recognition as reinforcement learning and has a fixed language policy without exploring the observer's strategy for asking clarification question.
\citet{holler2018plan} employs HTN Planning for PGR but struggles with sub-optimal actions and noisy sensors, focusing on handling missing sensor observations instead.
\nnote{Like \cite{holler2018plan} our work assumes that the goals are always achievable, they are unable to infer likely goals when irreversible failures occur.} 

\textbf{Context-aware Social Robotics:} \nnote{ First describe the work, then differentiate it from this work.}
In the past, research in social robotics has focused on developing non-verbal social behaviors for robots to assist the elderly ~\citep{fasola2013socially, kidd2008robots} during task completion. These works place less emphasis on incorporating language feedback/observations for user intent inference.  Further, the robot dialog policy (if involved) does not account for the environmental context, dialog context, and user modeling. Research in situated human-robot dialog
 by \citet{danbohus,thomason2019improving, idreesgrounding} 
grounds speech response in the environment but asks clarification questions heuristically, using rule-based/greedy approaches and without using a decision-theoretic framework. Such heuristics are prone to failure as the tasks get complex and the environment sensors become complex and noisy.

\textbf{POMDP-based Collaborative Dialog:}
Partially observable Markov decision processes (POMDPs) provide a rich framework for planning under uncertainty.
They excel in optimizing agents' actions over long horizons in complex environments despite incomplete state information from noisy sensors \cite{hoey2010automated}. 
\citet{young2013pomdp} and \citet{doshi2007efficient} built POMDP-based dialog systems. However, these only use language as observations, not world observations, for belief updates to choose actions with the highest reward accumulation. \citet{whitney2017reducing} fuses language and world observations for object fetching tasks but does not model the user's mental state during multi-step task completion. The closest work to ours is by \citet{ hoey2010automated}, featuring an engaging assistant that incorporates only world observations and not language observations for multi-step tasks. 
They can infer human actions and psychological states through hand and towel tracking but can not handle multiple and concurrent tasks and backtracking and are limited to hand washing. 
Research has recently focused on using reinforcement learning in a collaborative dialog for interactive task learning \citet{chai2018language}. However, these works require an existing dataset for offline learning, while our planning approach doesn't necessitate data collection or learning.
\iinote{improve wording}
\iinote{Remove: Further, they identify a set of questions for the agent to inquire about the state of the environment. However, their state factorization does not represent human plans and goals.}

\nnote{Our work extends these focusing on the agent maximizing the expected reward given the uncertainty in the user's mental model and incorporate both the world and language observations.}
\iinote{The solution to the POMDP determines a decision policy—that associates actions with specific belief states—that
optimizes long-term expected accumulated reward.}


\vspace{-1mm}
\section{Technical Approach}
 \begin{figure}[t]
    \centering
    \includegraphics[width=0.8\linewidth]{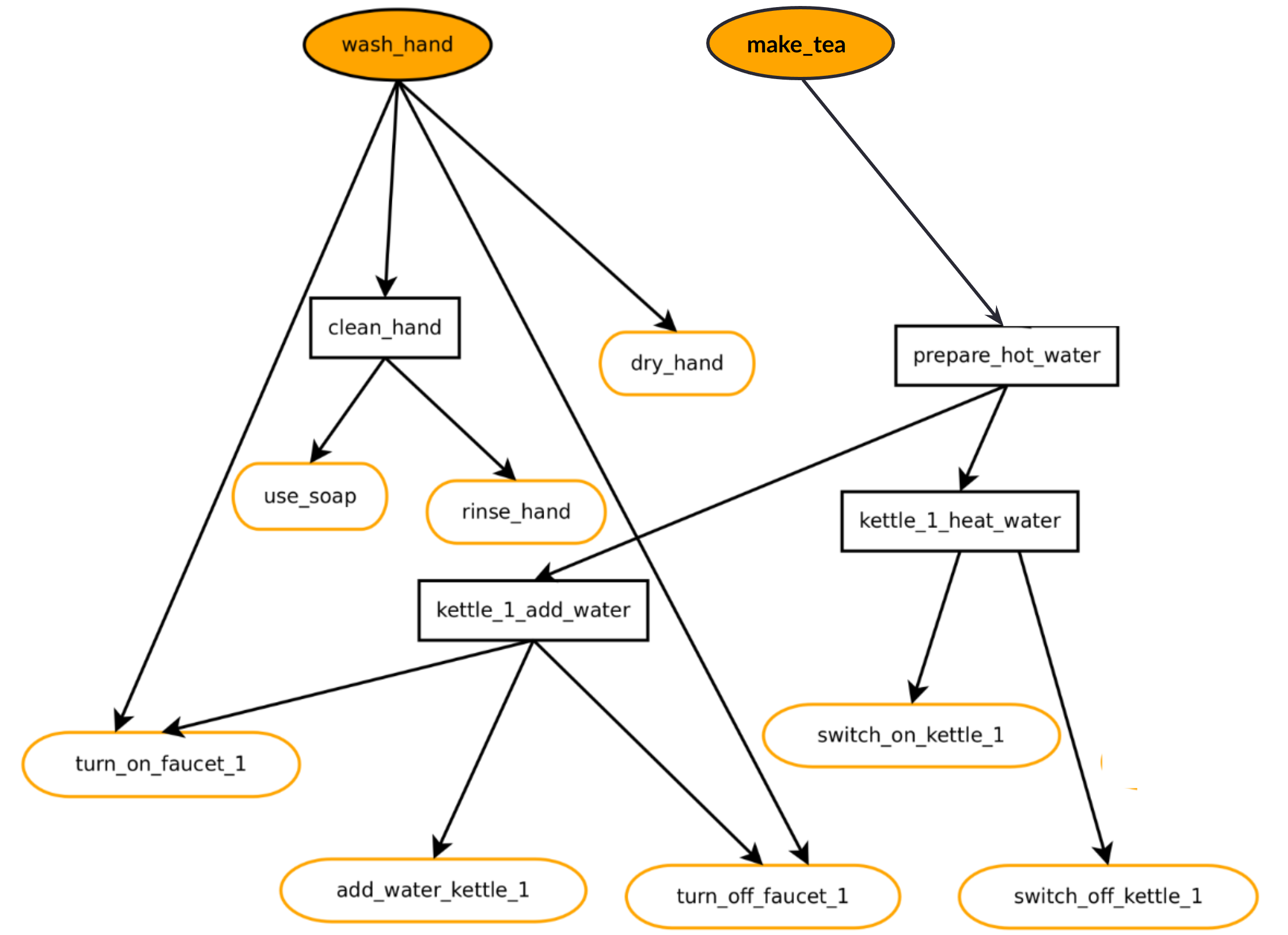}
    \caption{Hierarchical Task Network for two of our tasks -  washing hands, and making tea}
    \label{Fig-htn}
\end{figure}

Given the sensor measurements $O_w$ of the world state $W$, an assistive robot needs to infer the probabilities of the hidden user intent---the person's current goal $G$ and the current primitive human action $\alpha$. Partial observability of the latent user intent and noise in the sensor observations impact accurate inference of $G$, $W$, and $\alpha$  at every sequence of human actions.
We infer the probability distribution of the hidden states $G$, $W$, and $\alpha$ using HTN planning to generate feasible plans for the human-given predefined tasks similar to  work of \citet{wang2017hierarchical}.
\iinote{We do so by recursively decomposing the composite tasks utilizing a knowledge base and the sensor measurements at each time step to update our belief over human's progress.}   
 \nnote{The
algorithm adopts the hierarchical paradigm in Hierarchical Task Network (HTN) planning
[20], which generates feasible plans for predefined tasks by recursively decomposing composite tasks using a knowledge base. The modeling of ADLs with a HTN framework has
two advantages. First, the hierarchical nature of HTNs enables the algorithm to provide
prompts of different detail levels. Second, the knowledge base allows to set up specific
goals and preferences for the older adults.}
With the ability to ask clarification questions, the robot actively improves the inference of the latent states. The user's response in the form of language observations, $o_l$, gives the agent additional information about their task.
\iinote{We want the robot to decide when to ask questions when, and only when, it gets confused about the belief over human's progress while not bothering the user unnecessarily. We need to balance between information-gathering actions, like asking questions, and goal-inducing actions, such as providing the correct next step.}
With our D4GR framework, the robot decides when to ask clarification questions and which user action to inquire about, based on information gathered from both the sensors and language.
Our framework avoids asking unnecessary questions and balances between information-gathering actions, like asking questions, and goal-inducing actions, such as providing the correct next step. 

\subsection{POMDP Definition}
\nnote{Stefanie: this part goes in "Background," which is part o Section II - called it Background and Related Work. First, define your problem setting - what are the inputs and outputs of an interactive dialog setting? Then say you're going to model it as a POMDP - define a POMDP. Then talk about all the other systems. For Part III, only talk about the new things due to your work}

We model our PGR problem as a POMDP~\cite{kaelbling1998planning} planning problem,\finaliinote{ generating an approximately optimal action policy for the robot.} Formally, a POMDP is defined as a tuple ($S, A, T, R,\Omega, O,\gamma
,b_{0}$) where $S$ is the state space,  $A$ is a the action space, $T$ is the  transition probability, $R$ is the reward function,
$\Omega$ is a set of observations, $O$ defines an observation probability and  $\gamma$ is the discount factor. Since $s_t$ is not known exactly, the POMDP model updates, at each timestep, the probability distribution over all possible states (belief state $b_t$). The POMDP agent uses a planner to generate an optimal policy for the robot's action, which in this case is the communication with the human user.


\subsection{D4GR Formulation} \label{subsec: CD_formulation}
\begin{figure}[t]
    \centering
    \includegraphics[width=0.85\linewidth, keepaspectratio]{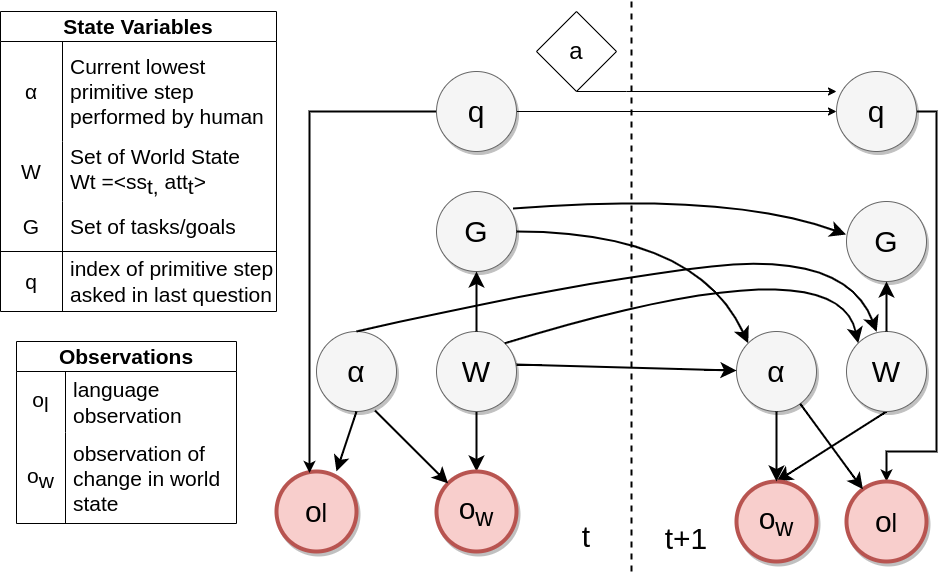}
    \caption{Influence diagram for D4GR.} 
    \label{Fig-Influence_diagram_coachdial}
    \end{figure} 

We define a novel model, \textbf{D}ialogue \textbf{4}for \textbf{G}oal \textbf{R}ecognition, a Partially Observable Markov Decision Process
(D4GR) that combines the goal and plan recognition components as described by  \citet{wang2017hierarchical} with the POMDP formalism to allow robots to take action in the environment through dialogue for improved PGR. 
For efficient human intent recognition and estimation of optimal action policy, our D4GR must
handle the large space of the world and language observations. Our formulation leverages the hierarchical task structure of HTN and assumes independent assumptions between state variables set of goals $G$, human user action $\alpha$, and world state $W$ for efficient belief update.
Our D4GR has the following components: ($S,A,T,R,\Omega,O,\gamma
;b_{0}$).

\textbf{State (S): } The state, $s_t \in S$, consists of a tuple of the user's mental state, $M_t$, and the world state, $W_t$, along with information needed to track the dialog state.   We represent the user mental state using HTNs. We assume access to the HTN's fixed knowledge graph $TaskNet$  for the tasks, where the root node(s) represent the high-level tasks/goals $G$ that the human can do. The internal nodes are sub-tasks that can be decomposed into leaf nodes depicting primitive human actions. $\alpha$ denotes the current primitive human action. We model the user's mental state $M$ represented by $G$ and $\alpha$. The partial $TaskNet$ for our kitchen domain is shown in Fig-\ref{Fig-htn}. 
\vspace{-1mm}

The world state $W_t$ combines the states of world smart sensors $ss_t$ and the attributes of objects involved in the task $att_t$, such as dryness of the hand, state of the faucet, etc. 
The dialog state variable $q_t$ stores the latest primitive human action referenced by the robot in its clarification question. Thus, the state $s_t$ can be factored into the following components:$s_t = (M_t, q_t,  W_t,)$
where $W_t$ = $(ss_t, att_t)$, $M_t=(G_t, \alpha_t)$. 
Here $G_{t}, \alpha_t, att_t, ss_{t}$ are the hidden variables, while $q_{t}$
is the known variables, hence making this a Mixed Observable Markov Decision Process (MOMDP) \cite{ong2009pomdps}.  The influence diagram can be seen in Fig-\ref{Fig-Influence_diagram_coachdial}.

\textbf{Action(A)} includes the actions of the agent. The robot for this research can perform the following predefined language-based actions - 1) $Wait$: does nothing but advances the time step.
2) $Ask\{argmax(\alpha)\}$:
 The robot asks a clarification question about the primitive action $\alpha$ with the highest belief. The question template used is: ``I believe that you just did action $(\alpha_{i})$, is this correct?''.
3) $inform\_next\_instruction$: informs the next action that the user should perform at the current timestep based on the current belief. This action is chosen based on a fixed policy and is executed when the user provides a negative language response to the robot's clarification question.

\iinote{inform\_next\_instruction} 
\textbf{Observations ($\Omega$)} encompass both the user's language ($o_l$) and observations about the world state ($o_w$). $o_{w}$ includes discrete observations of the world smart sensor's state $ss_t$ and the attributes of the task-related objects $att_t$ such as \emph{ hand\_dry == true, faucet\_on == false}, etc.
These observations are binary for the states in $W_t$, so the faucet can only be on or off. The language observations are natural language responses.

\textbf{Observational Model ($O$): } The robot needs a model of $p(o|s)$ $=$ $p(o_l, o_{w}|s)$ to update its belief. Most of the complexity of our model is captured in this observation model and belief update defined in the sec-\ref{sec-belief}.

\textbf{Transition Model ($T$) : }$T(s,a,s') \equiv p(s_{t+1}|s_t,a_t)$. 
Our stochastic transition function is factorized as shown in Eq-\ref{overall}, following a similar approach to \citet{wang2017hierarchical}. We factor our mental model $M_t$ into $G_t$ and $\alpha_t$. Additionally, we assume that the last question asked, $q_t$, is independent of $G$, $\alpha$, and $W$.
\begin{align*}
\tiny
p(s_{t+1}|s_t,a_t) 
&= p(G_{t+1} | W_{t+1},G_{t}) \times  p(W_{t+1} |  W_{t}, \alpha_{i,t+1}) \times\\
       & \qquad  p(\alpha_{t+1}|W_{t}, G_{t}) \times p(q_{t+1}|q_{t}, a). &\numberthis \label{overall} 
\end{align*}
In Eq-\ref{overall}, we assume that $q_t$ changes deterministically from null to $max{(\alpha_t)}$ after the robot asks a clarifying question. Further, $G$ is deterministically carried forward to the next time step.
\finaliinote{
\begin{align}
    &p(q_{i,t+1} | q_{i,t}, a_t) &\nonumber\\
\qquad = &\begin{cases}
    1 \text{ for } \tiny{max(\alpha)} \quad else  \quad 0,& \text{if } a \neq NULL\\
   1 \text{ for } q_{i,t},\quad else \quad  0              & a_t == NULL. \nonumber
\end{cases}&
\label{eq:question-transition}
\end{align}
}
\textbf{Reward $(R(s,a))$ }
    \nnote{Reward for goal, cost for asking clarification question and time to stay alive}
    We provide a positive reward (5) for asking a clarifying question when the user is doing the wrong or suboptimal lowest primitive step. A negative reward ($-5$) for asking a clarifying question when the human user is doing the correct primitive step or when the agent asks a question about a wrong primitive action. Thus, doing nothing accumulates zero rewards until the right question is asked, while not asking a question or asking a wrong one results in a penalty.
     \nnote{st: cut, this doesn't make sense with the +5 reward anyway:   The costs of the different actions were initially set to correspond to the number of seconds it would take to complete said action and were tuned from there using both simulated trials and our small pilot study.} 

    
\nnote{We assume that we have access to a knowledge graph for the tasks (HTN) with sub-tasks as nodes that can be decomposed into leaf nodes depicting primitive human actions. HTN for our work is shown in Fig-\ref{Fig-htn}.}

\nnote{\hspace{2mm} The current world state transition -Eq-\ref{eq:world-transition} and human action transition - Eq-\ref{eq:currstep-transition} are rooted in how $TaskNet$ works as a part of HTN. An internal/root node can be decomposed into its children subtasks node if the preconditions are satisfied, leading to stochastic effects in the children nodes.
This node decomposition process is defined 
in \citet{wang2017hierarchical} where the execution of a primitive human action $\alpha_{i,t}$n causes the predefined effects of $\alpha_{i,t}$ to be triggered in the successor state -  $w_{t+1}$
according to the successor node function of HTN predefined \citep{wang2017hierarchical}. The human primitive action transition - $p(\alpha_{i,t+1} | w_{i,t}, G_{i,t+1})$ -- Eq-\ref{eq:currstep-transition} and the constant C are adopted from the extensive experimented test cases introduced in \citet{wang2017hierarchical}.}


\subsection{Belief Update for Goal Recognition and Planning}\label{sec-belief}
\begin{figure*}[t]
    \centering    \includegraphics[width=\linewidth]{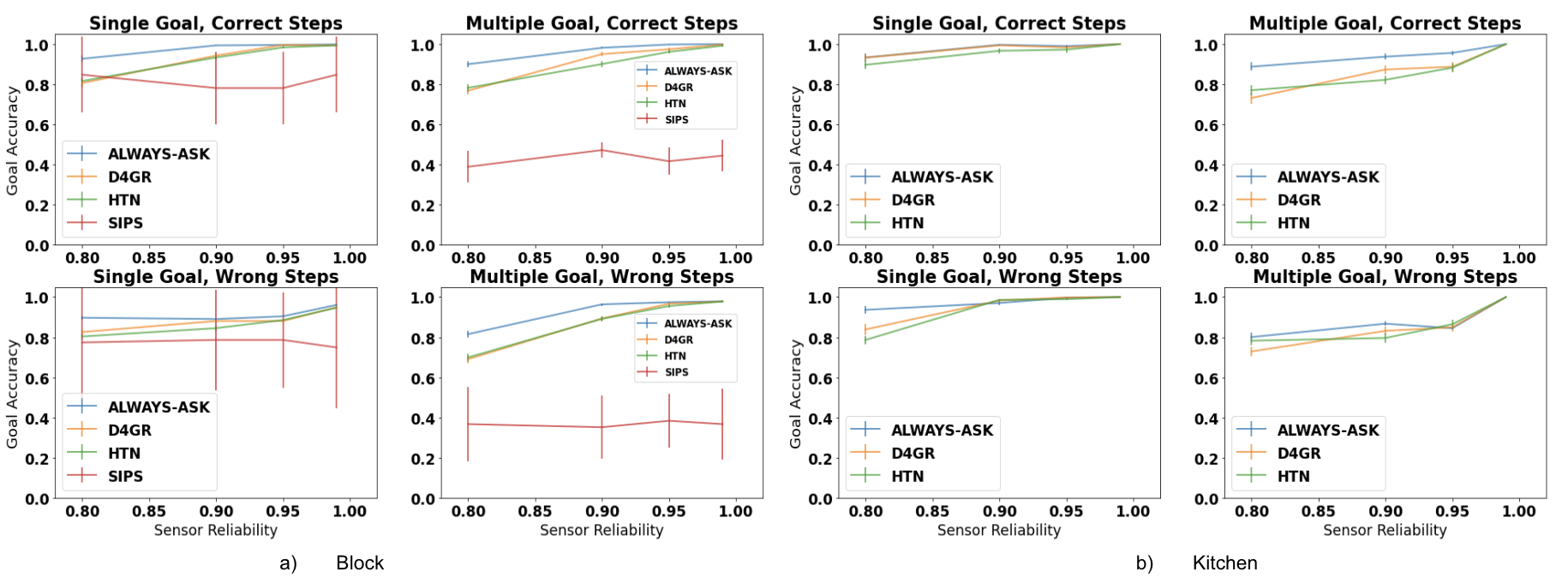}
    \caption{Results for Top1 Goal Accuracy versus Sensor Reliability for the two domains kitchen and block}
    \label{Fig-result_acc}
    \end{figure*}

Our belief update performs human intent recognition by maintaining a belief over the hidden user's mental state $M_t =(G_t, \alpha_t)$ and the world state $W_t$.
The actions executed by the user produce an observation of the world state $o_{w}$ indicating the change in the world state $W_t$. The user can also provide speech/language feedback $o_{l}$ in response to the clarification question asked. We classify the intent of each sentence into positive or negative feedback using the bag of words approach. Negative responses $r_n$ include \{ ‘no’, ‘nope’, ‘other’, ‘not’ \} while positive responses $r_p$ include the words \{ ‘yes’, ‘yeah’, ‘sure’, ‘yup’ \}.
Further, our world sensor noise model 
generates the correct sensor state with probability $sr$ and the incorrect sensor state with $1-sr$. We adopt the noise model for the sensor described by \citet{wang2017hierarchical}. 
\iinote{ As a result:
    \begin{equation} \label{o_exp}
    o_t=<o_{l,t}, o_{w,t}>, 
    \end{equation}
where $o_t\in \omega_t$}
\nnote{use summation to derive 8 from 6, we should pull language out, rephrase - convert joint distribution into summation and pull language out}
\iinote{\begin{align}
    & b'(s_t) = p(s_t|o_{t}, a_{t-1},b_{t-1}) &\nonumber\\
    & \quad\quad = n O(o_{t}, s_{t},a_{t-1}) \sum_s T(s_{t-1},a_{t-1},s_t)b(s_{t-1}). &\numberthis \label{eq-beliefupdate}
\end{align}}

The observation model can be further expanded and approximated as follows:
\begin{align}
\tiny
&p(o_t|s_t; a_{t-1})\propto p(s_t|o_t,a_{t-1}) * p(o_t|a_{t-1}). &\numberthis
\end{align}
\nnote{Overall the derivation of probablity $s_t$ given $o_t$ and $a_{t-1}$ is expanded eq-\ref{eq-observationstateupdate} to Eq-\ref{eq-factormodels}.}
\iinote{st: You again don't need to show every single step.  I like to write it all out to vet the math; once it's vetted you can skip steps when explaining it.}
\nnote{\begin{align}
    & p(s_t|o_t, a_{t-1})&\numberthis\label{eq-observationstateupdate}\\
    & \propto p(G_{t},  \alpha_t, W_t, q_t, o_{l,t}, o_{w,t}, a_{t-1} )&\label{eq-jointdist}\numberthis\\
    & \propto p(G_{t},  \alpha_t, W_t, q_t | o_{l,t}, o_{w,t}, a_{t-1}) *  p(o_{l,t}, o_{w,t}, a_{t-1}) &\label{eq-factor1}\numberthis\\
    & \propto p(G_{t}, W_t| o_{l,t}, o_{w,t}, a_{t-1}) * p(q_t | o_{l,t}, o_{w,t}, a_{t-1})&\nonumber\\
    &  \qquad p(\alpha_t | o_{l,t}, o_{w,t}, a_{t-1}) * p(o_{l,t}, o_{w,t}, a_{t-1}). &\label{eq-factor2}\numberthis
\end{align}}

\nnote{
\begin{align}
    & p(s_t|o_t, a_{t-1})&\numberthis\label{eq-observationstateupdate}\\
    & = p(G_{t},  \alpha_t, W_t, q_t | o_{l,t}, o_{w,t}, a_{t-1}) *  p(o_{l,t}, o_{w,t}, a_{t-1}) &\label{eq-factor1}
\end{align}}

Overall, the probability of $s_t$ given $o_t$ and $a_{t-1}$ can be factored into the  world observation model and language observation model in Eq-\ref{eq-factormodels}. We assume that the world observation $o_{w,t}$ solely provides information about the $W$ and $\alpha$. Meanwhile, the language observation is relevant to the human action $\alpha$, the last question asked $q_t$, subsequently affecting the goal. 
\begin{align}
\tiny
    & p(s_t|o_t, a_{t-1}) \propto  \underbrace{p(G_t|W_t)*p(W_t|o_{w,t})*p(\alpha_t| o_{w,t}) * p(o_{w,t})  \rule[-5pt]{0pt}{5pt}}_{\mbox{world observational model}} &\nonumber\\
    &* \underbrace{p(\alpha_t, q_t| o_{l,t}) * p(o_{l,t}) * p(q_t | a_{t-1}) \rule[-5pt]{0pt}{5pt}}_{\mbox{language observational model}}.&\label{eq-factormodels}\numberthis
\end{align}

\iinote{dealt with action in eq-\ref{eq-independence_ass1}?}\iinote{dealt with action in eq-\ref{eq-factormodels}?}

For both the world and primitive action belief update in eq-\ref{eq-factormodels}, the components $p(W_t|o_{w,t})$ and $p(\alpha_{t}| o_{w,t})$  are derived from \citet{wang2017hierarchical}. The Bayesian update is as follows:
\begin{align}
\tiny
     &p(\alpha_{t}| o_{w,t})
     \propto \sum_{w_{t-1} \in W_{t-1}} \sum_{w_{t} \in W_{t}} p(\alpha_t,  o_{w,t}, w_{t-1}, w_t),& \label{eq-currstep}\numberthis
\end{align}
\begin{align}
\tiny
    & p(W_t|o_{w,t})=\sum_{w_{t-1} \in W_{t-1}}  \sum_{\alpha_{i,t} \in \alpha_{t}} p(\alpha_{i,t}, o_{w,t}, w_{t-1}, w_t). & \label{eq-world}\numberthis 
\end{align}

\iinote{ Both human action recognition result $p(\alpha_{t}| o_{w,t})$ and world state update $p(W_t|o_{w,t})$ adopt Bayesian update as described below in Eq-\ref{eq-currstep} \nnote{fix this equation} and Eq-\ref{eq-world}.} 

We adopt the proposed algorithm for goal recognition, $p(G_t|W_t)$ in \citet{wang2017hierarchical}.
\iinote{that relies on explanations $expla$.} 
The algorithm maintains a goal belief distribution by generating a probabilistic explanation set - $ExplaSet$. Each $expla \in Explaset$ uses HTN planning to explain the observations so far.
The probability of each goal $g_i$ in $G$ given the world state is the sum of probabilities of $expla \in ExplaSet$  whose $PredictedGoal$ == $g_i$. 
Our algorithm reweighs primitive actions probabilities based on the language observational model described below, influencing the world belief update according to Eq-\ref{eq-world} and, consequently, the goal recognition update.

\iinote{The probability of goal $p(G_t|W_t)$ is the sum of probabilities of $expla \in ExplaSet$  whose $startGoal$ contains $g_{i,t}\in G_t $.}

\nnote{Appendix: Both step recognition result $PS'_{t,posterior}$ 
and $W_{t, posterior}$ 
\iinote{do I need to show more explicit connection between $att_{i,t}$ and $W_{t}$  }
which is updated according to Eq-\ref{eq:wt} are fed as input to update $ExplaSet$. This updated $ExplaSet$ is used to calculate Posterior of $G_t$. 
Given a step $ps_{i,t} \in PS'_{t,posterior}$, each explanation $expla{i,t} \in ExplaSet_{t}$ will be updated to several new ones, which are stored in $ExplaSet_{posterior}$.
This includes two procedures: recognition and decomposition. Details can be found in \cite{wang2017hierarchical}. The important equation to remember is that the  
probability of new explanation is computed using Equation-\ref{eq:ex}
\begin{align}
    &p(new_{expla})= p((ps_{i,t})_{posterior}) * goalNet(expandProb) & \nonumber\\
    &\qquad \qquad \qquad *p(expla) \label{eq:ex}& 
\end{align}
The probability of goal $g_{i,t}$ in $G_t$ is the sum of probabilities of $expla \in ExplaSet$  whose $startGoal$ contains $g_{i,t}$. The probability of $ (ps_{i,t})_{posterior} \in PS_t$ is updated to be the sum of probabilities of explanations $expla \in ExplaSet$ whose $forest$ contains a node standing for $ps_{i,t}$ with completeness being false while readiness being true.
}

\nnote{Prior state probability is calculated as follows:
\begin{align}
    & \text{ prior } = p(s_t) &\nonumber\\
    & \text{ where }  s_t =  <G_{t}, CurrStep_t,  ss_t, att_t, q_t> &\nonumber\\
    &\propto \sum_{a_{t-1} \in A_{t-1}} \sum_{s_{t-1} \in S_{t-1}} p(s_t, s_{t-1}, a_{t-1})  &\nonumber\\
     &\propto \sum_{a_{t-1} \in A_{t-1}} \sum_{s_{t-1} \in S_{t-1}} p(s_t| s_{t-1}, a_{t-1}) * p(s_{t-1}) * p(a_{t-1}))  &\numberthis
\end{align}}
\iinote{This assumption of cooperation is what allows the robot to gather more information from the speaker's utterance, making the interaction quicker and more efficient. }
The derivation of the language observational model is: 
\begin{align}
    & p(\alpha_t, q_t| o_{l,t}) * p(o_{l,t}) \propto p(o_{l,t} | \alpha_t, q_t). &\label{eq-langmodel}\numberthis
\end{align}

\finaliinote{We adopt a bag-of-words approach as our POMDP's observational model instead of utilizing a large language model (LLM) like GPT3. LLMs are not inherently grounded. Our model explicitly establishes a connection between sensor information and semantics through a transition model in the POMDP. Although LLMs could be incorporated for intent classification using the right prompt, we did not pursue this direction as it falls outside the focus of our paper.} 

To estimate the effect of the language observation $o_l$ on $\alpha$ and $q$, we calculate $p(o_{l,t} | \alpha_t, q_t)$. For this, we consider three possibilities for the state:
If the highest belief primitive action $\alpha_{max,t}$ is the same as the question asked,  then the user is likely to respond with positive/confirmation feedback. 
The opposite is true if $\alpha_{max,t} \neq q_t$. If $q_t = Null$, then no question has been asked, so both types of responses are equally likely. The mathematical representation of $p(o_{l,t} | \alpha_t, q_t)$. is governed by the following conditional probability table:

\begin{table}[h]
    \small
    \centering
    \scalebox{0.9}{
    \begin{tabular}{l c c}
    \toprule
      & $p(o_{l,t} = Yes)$& $p(o_{l,t} = No)$ \\
    \midrule
    $\alpha_{max,t} = q_t$ & $0.99$ & $0.01$\\
    $\alpha_{max,t} \neq q_t$ & $0.01$ & $0.99$\\
    $q_{t} = Null$ & $0.5$ & $0.5$\\
    \bottomrule
    \end{tabular}
    }
    \caption{Conditional Probability for $p( o_{l,t+1}| \alpha_{t}, q_{t})$}
    \vspace{-7mm}
    \label{tab:o==Yes}
\end{table}

At each time step as the human performs an action, we solve the MOMDP using the POUCT solver \cite{silver2010monte} to approximate the optimal policy for the robot's communication with the human. The observational model is then employed to update the robot's belief of user's mental state $M = \{G,\alpha\}$.

\nnote{
For language feedback, $p(G | G, q, o_l)$
\begin{align}
    & \text{language belief update = } p(CurrStep_t, q_t| o_{l,t})&\nonumber\\
    & \propto p(o_{l,t} | CurrStep_t, q_t, ) &\nonumber\\
\end{align}
\begin{align}
    & p(o_{l,t}  | q_{i,t}, currstep_{t}) = &\nonumber\\
& \begin{cases}
    p(o_{l,t}) * p(o_{l,t}  | q_{i,t}, currstep_{i,t}) ,& \text{if } o_{l,t} \neq NULL \\
    1 - o_{l,t} ,              & \text{otherwise}
\end{cases} &
\end{align}
}
\nnote{
\begin{align}
    & p(s|o) \propto p(G_{t},  CurrStep_t, W_t, q_t, o_{l,t}, o_{w,t} )&\nonumber\\
    & \sum_{w_{t-1} \in W_{t-1}} \sum_{currstep_{t} \in CurrStep_{t}} (p(CurrStep_t| o_{w,t})) &\nonumber\\
    &\propto \sum_{w_{t-1} \in W_{t-1}} \sum_{currstep_{t} \in CurrStep_{t}} \Bigg ( \sum_{w_{t-1} \in W_{t-1}}  \sum_{w_{t} \in W_{t}}   &\nonumber\\
    &p(G_{t},  currstep_t, w_t, q_t, o_{l,t}, o_{w,t}, w_t, w_{t-1}) \Bigg ) &\nonumber\\
    & p(G_{t},  CurrStep_t, w_t, q_t, o_{l,t}, w_{t-1}, o_{w,t} ) &\nonumber\\
    & \propto\sum_{w_{t-1} \in W_{t-1}}  \sum_{w_{t} \in W_{t}} &\nonumber\\
    & p(w_t| currstep_{t}, w_{t-1}) * p(o_{w,t} | w_t) * p(w_{t-1}) * p(currstep_{t}) &\nonumber\\
    &* p(G_{t}, q_t, w_t, o_{l,t}, o_{w,t})&\nonumber\\
    &  \propto p(G_{t},  CurrStep_t, ss_t, att_t, q_t, o_{l,t}, o_{w,t} )&\nonumber\\
    & = \frac{\splitfrac{ \underbrace{p(o_{l,t}, o_{w,t} | G_{t}, CurrStep_t, ss_t, att_t, q_t) \rule[-5pt]{0pt}{5pt}}_{\mbox{observational model}} } {* p( G_{t}, NS_t, CurrStep_t,ss_t, att_t, q_t)}}{p(o_{l,t}, o_{w,t})} & \numberthis
\end{align}
\begin{align}
    & p(s|o) = p(G_{t},  CurrStep_t, ss_t, att_t, q_t|o_{l,t}, o_{w,t} )&\nonumber\\
    & = \frac{\splitfrac{ \underbrace{p(o_{l,t}, o_{w,t} | G_{t}, CurrStep_t, ss_t, att_t, q_t) \rule[-5pt]{0pt}{5pt}}_{\mbox{observational model}} } {* p( G_{t}, NS_t, CurrStep_t,ss_t, att_t, q_t)}}{p(o_{l,t}, o_{w,t})} & \numberthis
\end{align}
\begin{align}
    &\text{observational model} = &\\
    &p(o_{l,t}, o_{w,t} | G_{t}, q_{t}, CurrStep_t,  ss_t, att_t) &\nonumber\\
    & = \underbrace{p(o_{l,t}| G_{t},CurrStep_t,  ss_t, att_t, q_t, lo_t) \rule[-5pt]{0pt}{5pt}}_{\mbox{language observational model}} * &\nonumber\\
    & \underbrace{p(o_{w,t} | G_{t}, CurrStep_t,  ss_t, att_t, q_t, lo_t) \rule[-5pt]{0pt}{5pt}}_{\mbox{world observational model}}  &\numberthis
\end{align}
\bknote{Question: Is the following step okay?}\\
if $o_{w,t} \neq Null$ then posterior of state is as follows:
\begin{align}
    & \text{world observational model = } &\\
    & p(o_{w,t} | G_{t}, CurrStep_t,  ss_t, att_t, q_t)&\nonumber\\
    & = \frac{\splitfrac{ \underbrace{p(G_{t}, CurrStep_t,  ss_t, att_t, q_t|o_{w,t}) \rule[-5pt]{0pt}{5pt}}_{\mbox{state posterior}} } {* p(G_{t}, CurrStep_t,  ss_t, att_t, q_t)}}{p(o_{w,t})}&\numberthis
\end{align}
\bknote{does the following makes sense?, what to do with $G_t$ and its relation to $o_t$ and parent of ow is W}
\begin{align}
    & \text{state posterior} = &\nonumber\\
    &p(G_{t}, CurrStep_t, ss_t, att_t, q_t|o_{w,t})&\nonumber\\
    & \propto p(G_{t}, CurrStep_t, ss_t, att_t, q_t, o_{w,t})  &\nonumber\\
    & \propto P(CurrStep_t|o_{w,t}) * P(att_{t}|o_{w,t}) * p(G_{t} | o_{w,t}) &\nonumber\\
    &* P(ss_t|o_{w,t})* P(q_t|o_{w,t}) &\nonumber\\
    & \propto P(CurrStep_t|o_{w,t}) * P(att_{i,t}|o_{w,t}) * p(G_{t} | o_{w,t}) 
    \label{eq-posterior_state_world}
\end{align}
The terms in Eq-\ref{eq-posterior_state_world} is calculated as described by \cite{wang2017hierarchical}. It is also mentioned in our appendix section.
\bknote{Question -  Ask Eric does this makes sense?}
if $o_{w,t} == Null$ then posterior of state is as follows:
\begin{align}
    &p(o_{w,t} | G_{t}, q_{t}, CurrStep_t,  ss_t, att_t)&\nonumber\\
    &=1-p(o_{w,t})& 
\end{align}
where $p(o_{w,t}) = 0.95$ \\
}

\nnote{\appendix
The belief update a time step $t$, given $o_{w,t}$ is as follows. Posterior of $ps_{i,t} \in PS_t$ is calculated by the method adopted in HTN using Equation-\ref{eq:posterior_ps_t}. 
\begin{align}
    & (ps_{i,t})_{posterior} = p(ps_{i,t} | o_{w,t}) &\nonumber\\
    &= \frac{p( ps_{i,t}, o_{w,t})}{p(o_{w,t})}  \propto p( ps_{i,t}, o_{w,t}) &\nonumber\\
    &=\sum_{w_{i,t}} \sum_{w_{i,t-1}}  p( ps_{i,t}, o_{w,t}, w_{i,t}, w_{i,t-1}) &\nonumber\\
    &= \sum_{w_{i,t}} \sum_{w_{i,t-1}} p(w_{i,t}| w_{i,t-1}, ps_{i,t}) * p(o_{w,t} | w_{i.t})&\nonumber\\
    &  \qquad  \qquad \qquad \qquad \qquad  * p(w_{i,t-1}) * p(ps_{i,t})  &\numberthis \label{eq:posterior_ps_t}
\end{align}

$ss_t \subset W_t$ is fully observable and Posterior of $att_{i,t} \in att_t \subset W_t$ is dependant on $(ps_{i,t})_{posterior}$ and is calculated as follows:

\begin{align}
    & (att_{i,t})_{posterior} = p(att_{i,t} | o_{w,t}) &\nonumber\\
    &=\sum_{att_{i,t-1}} \sum_{ps'_{i,t} \in (ps_{i,t})_{posterior}}  p(att_{i,t}| att_{i,t-1}, ps'_{i,t}) * p(o_{w,t} | att_{i,t}) &\nonumber\\
    & \qquad  \qquad \qquad \qquad \qquad * p(w_{i,t-1}) * p(ps'_{i,t}) &\label{eq:wt} 
\end{align}
}



\section{Evaluation}
\begin{figure*}[t]
\centering    \includegraphics[width=\linewidth]{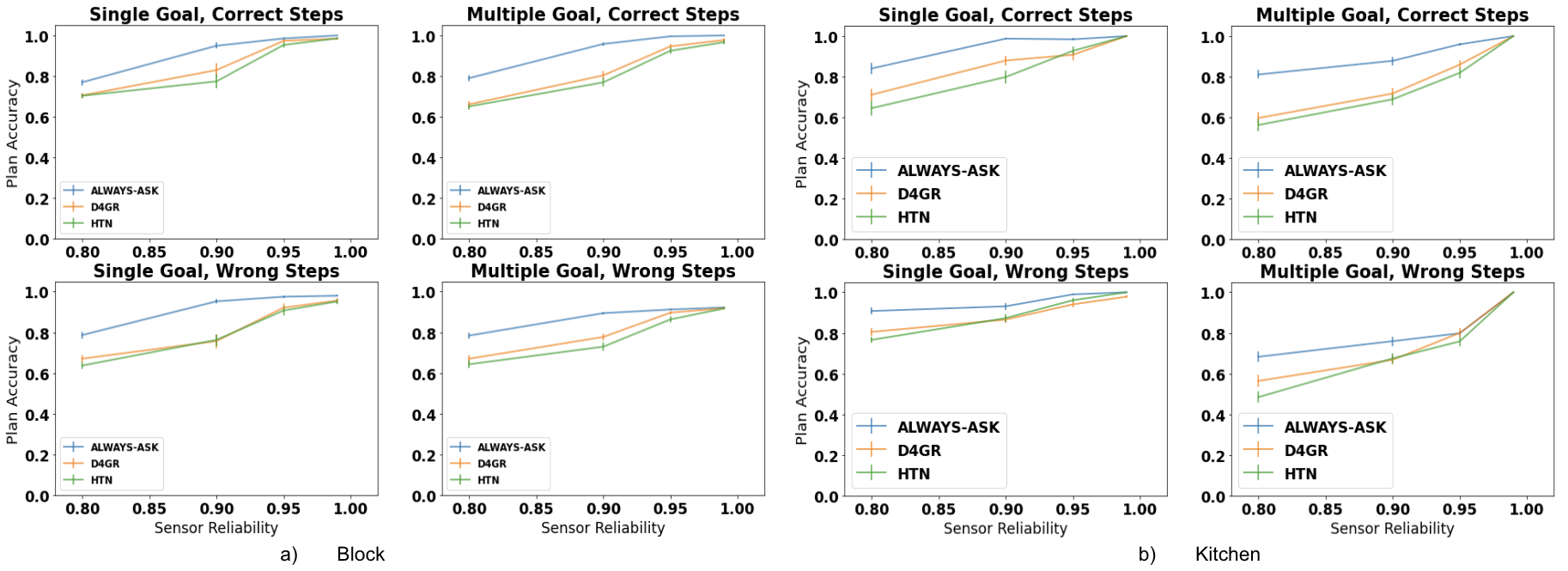}
\caption{Results for Top1 Plan Accuracy versus Sensor Reliability for the two domains kitchen and block}
\label{Fig-pa}
\vspace{3mm}
\end{figure*}

Our evaluation aims to test the hypothesis that our hierarchical decision-theoretic framework D4GR improves 1) the accuracy of goal recognition and plan recognition of human activity and 2) the robot's ability to guide the person towards task success.
\nnote{the expected return in guiding the user in completing the tasks successfully by observing the human actions and strategically asking clarification questions and instructing the human when the feedback is opposite to robot's own belief. }
We evaluate the enhanced performance of our algorithm by measuring the accuracy of goal recognition and prediction of the next human action, also referred to as plan recognition, at every time step. We also measure the planning time, the cumulative expected return, and the number of clarification questions asked for completing the tasks by D4GR and compare it against the three presented methods in simulation. We also perform a robot demonstration of D4GR for the scenario where the human switches between washing hand and making tea tasks. 

We use the simulation environment introduced by \citet{wang2017hierarchical} for our experiments. \finaliinote{The simulator models real environment state changes that result from the primitive actions specified in the HTN for the virtual human.} In the simulator, 44 binary virtual sensors are observing the world state. Some of the examples include sensors for $hand\_dry, faucet\_on, block\_picked\_up$. For our experiments, we vary sensor reliability from 99\% to 80\%.
\iinote{The list of all the sensors is added in the supplementary section}


\subsection{Domain and Experiment Test Cases}
We test our algorithm in two domains: a block domain and a kitchen domain. The Knowledge Base, $TaskNet$ of the HTN for the kitchen has three goals: \textit{wash hands}, \textit{make tea}, and \textit{make coffee}. 
and the block domain has five goals of stacking blocks to make words with varying lengths of 4 to 7. The five goals of our block domain are $rote$, $tone$, $tune$, $hawk$, and $capstone$. \finaliinote{
The two domains differ in their HTN planning structure, as the block's domain has higher goals (root nodes) but a shorter tree depth than the kitchen domain. Such a setup allows us to explore the effect of HTN structure on goal and plan recognition performance.}
We evaluate the performance of our algorithm over four categories of test case scenarios:

\textbf{Single Goal \&  Correct Steps:} Captures scenarios where the human always executes a correct action sequence for achieving a single goal. 

\textbf{Multiple Goals \&  Correct Steps:}  
The person works on multiple goals simultaneously by switching back and forth,

\textbf{Single Goal \& Wrong Steps:} A human has a single goal but can execute wrong steps affecting progress toward the goal. 

\textbf{Multiple Goals \&  Wrong Steps:} The human moves back and forth between goals and executes wrong actions. 

The easiest case is Single Goal \&  Correct Steps, and the hardest is Multiple Goals \&  Wrong Steps.
\subsection{Baselines:} We compare D4GR's performance in simulation with three other methods.  Our first baseline is \textbf{HTN}, a previous method of HTN-based goal recognition introduced by \citet{wang2017hierarchical}. This method passively incorporates partially observable world observations for PGR without engaging with the user. 
Our second method is \textbf{ALWAYS-ASK} 
which acts as an oracle that always asks the correct clarification question and uses the language feedback for the belief update of D4GR.
This baseline always has the highest goal and step recognition accuracy but receives a lower reward because it asks unnecessary questions.
\iinote{Our second baseline is a \textbf{ALWAYS-RANDOM} baseline that uses the belief update of DGR-POMDP but instead of performing planning to decide when to ask a clarification question, the agent always asks a  random question. Randomness in the clarification question is added by randomly choosing the human action parameter to fill the question template - ``I believe that you just did action $(\alpha_{i})$, is this correct?''  
Randomness in the question is added by randomly choosing the human action parameter to fill the question template  ``I believe that you just did action $(\alpha_{i})$, is this correct?'' } 
Our third baseline is \textbf{SIPS}\footnote{\finaliinote{The SIPS baseline 
is adopted from the code repository cited in \citet{zhi2020online}
and uses their default setting: static goal transition.}} introduced by \citet{zhi2020online}. 
This algorithm is not equipped to handle 
hierarchical nature of goals.

\subsection{Metric Definitions:}

\iinote{Tian add about performance metric from wang 
page 78}
We measure 1) Top 1 Accuracy for Goal recognition, 2) Accuracy for Plan Recognition similar to \citet{wang2017hierarchical} averaged over all timesteps. These metrics measure the accuracy of our belief update. To evaluate our POMDP formulation, we also measure the planning time taken: runtime averaged over the steps, the cumulative reward for the whole human action sequence, and the number of questions asked averaged over trials.



\section{Results}
Our proposed algorithm aims to improve the capability of the robot for goal and plan recognition. The performance of D4GR depends on how accurately our cognitive assistive robot estimates the belief states for the human mental model $M_t$: the likelihood of goals $G$ and the human actions $\alpha$ at each simulated step. The ground truth of each human action's $\alpha$ given the goal $G$ can be obtained from the knowledge base. 

\subsection{Exp 1 - Goal Accuracy Performance}\label{1}


\begin{table*}[t]
\begin{tabular}{|ll|rrrr|llll|llll|llllllll}
\cline{1-14}
                                                         &                     & \multicolumn{4}{l|}{\textbf{Runtime (s)} $\downarrow$ }                                                                                                                                 & \multicolumn{4}{l|}{\textbf{Cumulative Expected Return} $\uparrow$}                                                                                                                                                                                                   & \multicolumn{4}{l|}{\textbf{ Question Frequency} $\downarrow$}                                                                                                                                                                                                        &  &  &  &  &  &  &  &  \\ \cline{1-14}
\multicolumn{1}{|l|}{\textbf{Domain}}                    & \textbf{Method}     & \multicolumn{1}{l|}{\textbf{0.8}}       & \multicolumn{1}{l|}{\textbf{0.9}}       & \multicolumn{1}{l|}{\textbf{0.95}}      & \multicolumn{1}{l|}{\textbf{0.99}}      & \multicolumn{1}{l|}{\textbf{0.8}}        & \multicolumn{1}{l|}{\textbf{0.9}}                            & \multicolumn{1}{l|}{\textbf{0.95}}                           & \textbf{0.99}                                                 & \multicolumn{1}{l|}{\textbf{0.8}}                          & \multicolumn{1}{l|}{\textbf{0.9}}                          & \multicolumn{1}{l|}{\textbf{0.95}}                         & \textbf{0.99}                                               &  &  &  &  &  &  &  &  \\ \cline{1-14}
\multicolumn{1}{|l|}{}                                   & \textbf{ALWAYS-ASK} & 2.085                                   & 1.869                                   & 1.563                                   & 1.367                                   & \cellcolor[HTML]{FFFFFF}-37.253          & \cellcolor[HTML]{FFFFFF}-62.769                              & \cellcolor[HTML]{FFFFFF}-69.323                              & \cellcolor[HTML]{FFFFFF}-73.569                               & \cellcolor[HTML]{FFFFFF}1.000                              & \cellcolor[HTML]{FFFFFF}1.000                              & \cellcolor[HTML]{FFFFFF}1.000                              & \cellcolor[HTML]{FFFFFF}1.000                               &  &  &  &  &  &  &  &  \\
\multicolumn{1}{|l|}{}                                   & \textbf{D4GR}       & \cellcolor[HTML]{FFFFFF}\textbf{28.750} & \cellcolor[HTML]{FFFFFF}\textbf{20.445} & \cellcolor[HTML]{FFFFFF}\textbf{18.814} & \cellcolor[HTML]{FFFFFF}\textbf{17.084} & \cellcolor[HTML]{FFFFFF}\textbf{-26.056} & \multicolumn{1}{r}{\cellcolor[HTML]{FFFFFF}\textbf{-24.656}} & \multicolumn{1}{r}{\cellcolor[HTML]{FFFFFF}\textbf{-23.014}} & \multicolumn{1}{r|}{\cellcolor[HTML]{FFFFFF}\textbf{-23.989}} & \multicolumn{1}{r}{\cellcolor[HTML]{FFFFFF}\textbf{0.297}} & \multicolumn{1}{r}{\cellcolor[HTML]{FFFFFF}\textbf{0.318}} & \multicolumn{1}{r}{\cellcolor[HTML]{FFFFFF}\textbf{0.325}} & \multicolumn{1}{r|}{\cellcolor[HTML]{FFFFFF}\textbf{0.358}} &  &  &  &  &  &  &  &  \\
\multicolumn{1}{|l|}{}                                   & \textbf{HTN}        & 9.056                                   & 1.524                                   & 1.371                                   & 1.040                                   & \cellcolor[HTML]{C0C0C0}                 & \cellcolor[HTML]{C0C0C0}                                     & \cellcolor[HTML]{C0C0C0}                                     & \cellcolor[HTML]{C0C0C0}                                      & \cellcolor[HTML]{FFFFFF}0.000                              & \cellcolor[HTML]{FFFFFF}0.000                              & \cellcolor[HTML]{FFFFFF}0.000                              & \cellcolor[HTML]{FFFFFF}0.000                               &  &  &  &  &  &  &  &  \\
\multicolumn{1}{|l|}{\multirow{-4}{*}{\textbf{Block}}}   & \textbf{SIPS}       & \cellcolor[HTML]{FFFFFF}43.173          & \cellcolor[HTML]{FFFFFF}41.615          & \cellcolor[HTML]{FFFFFF}40.654          & \cellcolor[HTML]{FFFFFF}40.934          & \cellcolor[HTML]{C0C0C0}                 & \cellcolor[HTML]{C0C0C0}                                     & \cellcolor[HTML]{C0C0C0}                                     & \cellcolor[HTML]{C0C0C0}                                      & \cellcolor[HTML]{C0C0C0}                                   & \cellcolor[HTML]{C0C0C0}                                   & \cellcolor[HTML]{C0C0C0}                                   & \cellcolor[HTML]{C0C0C0}                                    &  &  &  &  &  &  &  &  \\ \cline{1-14}
\multicolumn{1}{|l|}{}                                   & \textbf{ALWAYS-ASK} & 0.484                                   & 0.425                                   & 0.407                                   & 0.343                                   & \multicolumn{1}{r}{-41.848}              & \multicolumn{1}{r}{-55.593}                                  & \multicolumn{1}{r}{-62.342}                                  & \multicolumn{1}{r|}{-73.265}                                  & \multicolumn{1}{r}{1.000}                                  & \multicolumn{1}{r}{1.000}                                  & \multicolumn{1}{r}{1.000}                                  & \multicolumn{1}{r|}{1.000}                                  &  &  &  &  &  &  &  &  \\
\multicolumn{1}{|l|}{}                                   & \textbf{D4GR}       & \textbf{13.912}                         & \textbf{13.439}                         & \textbf{13.572}                         & \textbf{11.384}                         & \multicolumn{1}{r}{\textbf{-24.527}}     & \multicolumn{1}{r}{\textbf{-26.542}}                         & \multicolumn{1}{r}{\textbf{-26.399}}                         & \multicolumn{1}{r|}{\textbf{-21.437}}                         & \multicolumn{1}{r}{\textbf{0.312}}                         & \multicolumn{1}{r}{\textbf{0.317}}                         & \multicolumn{1}{r}{\textbf{0.316}}                         & \multicolumn{1}{r|}{\textbf{0.296}}                         &  &  &  &  &  &  &  &  \\
\multicolumn{1}{|l|}{\multirow{-3}{*}{\textbf{Kitchen}}} & \textbf{HTN}        & 0.415                                   & 0.380                                   & 0.391                                   & 0.349                                   & \cellcolor[HTML]{C0C0C0}                 & \cellcolor[HTML]{C0C0C0}                                     & \cellcolor[HTML]{C0C0C0}                                     & \cellcolor[HTML]{C0C0C0}                                      & \multicolumn{1}{r}{0.000}                                  & \multicolumn{1}{r}{0.000}                                  & \multicolumn{1}{r}{0.000}                                  & \multicolumn{1}{r|}{0.000}                                  &  &  &  &  &  &  &  &  \\ \cline{1-14}
\end{tabular}
\label{table-question}
\caption{Trend in Questions Asked, Rewards Accumulated and Runtime (Sensor reliability varies from 80\% to 99\%)}
\end{table*}
In Fig-\ref{Fig-result_acc}, we present results for the average goal accuracies of D4GR and compare them with the baselines over varying sensor reliability and test case categories.
The reason for choosing the sensor reliability range from [0.8 to 0.99] is because most of the deep-learned vision and human action detectors have similar average accuracy \cite{pareek2021survey}. Overall, as the sensor reliability decreases, the accuracy performance of HTN-based methods (ALWAYS-ASK, D4GR, HTN) suffers. The oracle baseline, ALWAYS-ASK, always has the highest goal accuracy. Even at lower sensor reliabilities (higher sensor noise), D4GR's accuracy remains higher than HTN in all experiment categories by 1\% on average in both domains.
This trend indicates that even when the sensor's observational model fails, D4GR can better predict the belief states than HTN. The SIPS method did not generate functional plans for our kitchen domain even when the input specification was correct. Their particle filter algorithm could not find feasible plans for the goals. 
Hence, presenting the results in the blocks domain.
Compared to the SIPS baseline, our method is 30 \% better in the blocks domain. We significantly outperform the SIPS baseline in the multiple goals scenario by almost 43\% because SIPS does not handle the hierarchical nature of goals.
The problem categories with single goals (correct steps \& wrong steps) have the best performance for the lowest sensor reliability. We see D4GR performance boost by 2.8\% in the kitchen domain and 1.4\% in the blocks domain as compared to HTN. The Oracle, on average, is 6.3\% more accurate than the HTN baseline in this category. Our D4GR improves accuracy by inferring when and what to ask a question rather than always asking.
\iinote{The results for HTN  presented in this paper are slightly different than those cited in \cite{wang2017hierarchical}. We generate the baseline results by running the codebase provided by the author on python 3.10. These results are  qualitatively similar but not identical. }
\iinote{The most challenging scenarios are where the sensor reliabilities < 0.8. In such scenarios, our algorithm is 9\% more accurate than the HTN baseline. The ALWAYS-ASK baseline is 6.2\% more accurate than HTN in all the trials. Our DGR-POMDP performs similarly to it especially at $sr<0.8$. }


\iinote{The performances positively correlate with
sensor reliabilities. When sensor reliabilities reduce, the average accuracies deteriorate as well. (2)
The easiest problem category, p1, targets problems
with a single goal and correct steps and has the best performance. The average accuracies are very high even when sensor reliabilities are only 0.8. (3) The hardest
problem category, p4, targets problems with multiple goals and wrong
steps and has the worst performance. The accuracies are acceptable only when sensor reliabilities are above 0.95. This result is reasonable since the algorithm deals with noisy sensors, multiple goals, and wrong steps. (4) The other two
categories, p2 and p3, have similar performances, which are acceptable when
sensor reliabilities are above 0.9.}

\subsection{Exp 2 - Plan Accuracy Performance}\label{2}

Similar to goal accuracy, we plot the planning accuracy for D4GR and compare it with the baselines for varying sensor noise in Fig-\ref{Fig-pa}. Our algorithm overall is  3\% more accurate than HTN in both domains. For the lowest sensor reliabilities, D4GR is 4\% better than HTN in the kitchen domain and 2\% better than HTN in the block domain. For the multiple goal scenarios (correct and wrong steps), D4GR performs the best with an accuracy improvement of 2.7\% in the kitchen domain and 2.3\% in the block domain.

\subsection{Exp 3 - Trend in Questions Asked, Rewards Accumulated and Runtime:} 

Our proposed algorithm aims to improve the PGR capabilities of the agent by enabling the robot to engage with the users and ask for language feedback.
One performance measure is the number of helpful clarification questions asked with varying sensor reliabilities in Table-II.
When the environment and human actions are unambiguous (sensor reliability is high and/or the human is performing correct actions), D4GR enables the robot to intelligently infer that it does not need to ask lots of questions. 
At sensor reliability 0.99, D4GR still asks questions because users can do sub-optimal actions leading to ambiguity. 
Overall the agent asks questions 32.4\% and 31\% of the time in the block and kitchen domains respectively. 
When compared over varying sensor noise, the change in the number of questions asked is insignificant; the numbers lie within the same standard deviation. 

Further, asking a large number of clarification questions, especially if they are not relevant to the current progress of the task, takes more computing resources since  planning will have to be done at every timestep. This effect is measured by the runtime and the reward accumulated by D4GR.
Our reward function penalizes asking a lot of questions, especially when they reference an irrelevant human action. Our results show that D4GR takes more planning time per step as compared to ALWAYS-ASK and HTN but is 48\% faster than SIPS. All these measurements are done on a machine with 31 GB RAM, Intel® Core™ i7-9750H CPU @ 2.60GHz × 12. The code was run single-threaded. D4GR takes more time than HTN but enables accuracy performance boost as noted in previous sections, Sec-\ref{1} and Section-\ref{2}. Further, D4GR accumulates a higher reward/lower penalty as compared to ALWAYS-ASK by 58\% in both domains highlighting that D4GR does not ask unnecessary questions. 

\iinote{\textbf{Results:} The average number of questions asked by DGR-POMDP with varying reliabilities are measured. Results are shown in in Fig-\ref{Fig-result_question}. As the sensor reliability decreases, our algorithm intelligently infers more clarification questions as the environment and the human action are more ambiguous. These clarification questions are still 66\%  smaller than the questions asked by the agent with a fixed policy of always asking a question. When the human is doing wrong step in p2 and p4, the rate of clarification's asked remains constant showing that our agent has learned a policy which is not just sensor noise sensitive but is also sensitive to human action. }
\iinote{Out of these questions, the rate of questions at incorrect times are:}

\subsection{Robot Demonstration}
We performed a robot demonstration to highlight the feasibility of D4GR in the real world. The Kuri robot was used primarily for the demo due to its audio transcription capability. The demonstration consisted of a human performing two interleaving tasks in a kitchen while the Kuri robot observed the actions performed and engaged with the user using the D4GR algorithm. The human begins with making coffee and then moves to wash their hands. 
\finaliinote{Unlike HTN, which struggled to recognize the change in goal, D4GR correctly identified both goals. 
The demonstration simulated sensor reliability at 0.8 using an oracle and sensor noise model.}  Kuri used D4GR to intelligently infer when to ask a question and what to ask and is able to perform goal recognition and planning correctly. In the case of negative feedback from humans, Kuri offers its predicted correct step to the human. 
\section{Discussion and Future Work}
Our deployed algorithm D4GR shows improved accuracy for goal and plan recognition than the baseline $HTN$ and $SIPS$. It does so while asking fewer questions than the ALWAYS-ASK oracle policy. 
\finaliinote{Our deployed robot with D4GR performs real-time communication, as demonstrated. The time taken for online planning is influenced by two critical parameters of the POMDP solver: 1) $d$, representing the finite depth of the probabilistic decision tree constructed with state-action pairs, and 2) $n$, the finite number of observations sampled from each node.  Increasing $d$ and $n$ enhances the solver's accuracy and increases runtime. To achieve real-time communication, we conducted empirical experiments and determined that setting $d = 19$ and $n = 6$ provided appropriate action choices within a reasonable time.}

\nnote{Our work is also dependent on the HTN representation that enables the algorithm to deal with ordered / unordered sub-tasks and alternative ways to achieve a goal. However, in the case of tasks with many branches and sub-branches, the goal recognition algorithm will have to explore every branch to build an $explaset$, increasing the time for online inference. Multiple steps executed by a human at time $t$ are not considered in this work, but the human can do numerous tasks in one session. }
\nnote{One thing that we noticed in some of the trial runs for our online inference was that if the first executed human action shared between multiple goals goes wrong, the baseline HTN algorithm goes into an irrecoverable state that cannot be recovered even with our language feedback. Further, it will also be interesting to extend the belief function to accommodate the scenario where the human is not rational and might provide wrong feedback.}

\finaliinote{
Our algorithm shows promising improvements in PGR accuracy, although it comes with increased runtime compared to HTN. Our algorithm is designed to assist users with cognitive impairment in their daily tasks, focusing on non-time-critical activities. By providing delayed feedback, our social assistive robot increases the likelihood of users learning from mistakes and avoiding continuous repetition of errors compared to using HTN. We can reduce runtime further by retaining only the highest probable explanation sets, denoted as $ExplaSet$ in HTN planning. However, this impacts the PGR accuracy since $ExplaSet$ with multiple goals during initial steps can get pruned due to their lower probabilities.
In the multi-goal and low sensor reliability setting, D4GR shows slightly lower PGR accuracy than HTN. This is due to D4GR's reliance on noisy beliefs and user switching goals, leading to a higher probability of asking questions about irrelevant actions. The rational language feedback also adversely affects the update of the explaset, potentially diminishing its utility in later timesteps of the episode.}

Our work is also limited by the type of clarification questions the robot can ask. We have a fixed template for the question. It will be interesting to see how humans respond to various clarification strategies and how the robot can plan over a space of such categories. This will increase the action space requiring  more exploration by the POMDP solver. Further, our language observational model is a bag of words model. It can be more expressive by incorporating inference from LLMs.

Further, our work assumes access to a pre-defined knowledge base for the tasks. One thing that we will be exploring in the future is how to make the knowledge base adaptive to a layman user's needs and preferences as the task progresses through interactive dialogue. Our research opens venues for language grounding and human intent recognition in other collaborative tasks like building machines/complex furniture together by humans and robots.  This is an encouraging step toward enhancing the sensory capabilities of home-service robots that can assist people in completing tasks with language-based interactions.






\nnote{The later paragraphs should cover future work. This part of the paper can recognize limitations of your paper and sketch out open problems that remain to be solved.It is okay if you do not plan to solve them; you are also laying out boundaries for what this paper does not solve, which helps people understand what it does solve. Moreover you are also describing possible follow-on work that may inspire someone to expand on this paper.}
\nnote{Multiple steps executed by a human at time $t$ are not considered in this work but the human can do multiple tasks in one session.
If the first step that the human is doing goes wrong then HTN cannot recover and so can't we.\\}
\nnote{
Questions from Research Symposium:
- Can the knowledge base be changed as the robot observes the user completing tasks?\\
- Can HTN extend to more complex tasks?\\
- Will the model still be able to perform accurately in a task with many branches and subtasks?\\
- What if the user gives the wrong feedback to the robot?\\
- There were also a lot of questions about how the robot would detect task completion, (ex. cameras, sensors, computer vision model)}
\section{Conclusion}
We propose a novel algorithm for robots  to interactively keep track of people's ongoing progress in a task using questions. Moreover, our D4GR framework can suggest plan improvements to users in solving a task if required.
Our work shows that: 1) modeling the user as an HTN and incorporating language feedback improves robots' belief of human's progress in simulation; 2) POMDPs are effective methods for tracking a task's progress and asking clarification questions. Our D4GR formulation has a similar goal and step recognition accuracy as the best baseline ALWAYS-ASK method while asking 68\% fewer questions. In future work, we aim to conduct a user study with the targeted population to measure our approach's usefulness during the interaction. D4GR's ability to intelligently balance between clarifying uncertainty with a lesser number of questions allows for realistic interactions between a social robot and a human. This ability in the future can allow for realistic interactions with human users during collaborations over tasks between humans and robots. 

\section{Acknowledgement}
\finaliinote{We thank our labmates at Brown and Rutgers for their valuable insights. This work was supported by NSF under award number IIS-1652561, ONR under award numbers N00014-21-1-2584 and N00014-22-1-2592, and with funding from Echo Labs.
}

\bibliographystyle{plainnat}
\footnotesize\bibliography{references}

\end{document}